\documentclass[letterpaper, 10 pt, conference]{ieeeconf}  %

\IEEEoverridecommandlockouts                              %

\usepackage{amsmath}
\usepackage{hyperref}
\usepackage{tikz}
\usetikzlibrary{shapes,arrows, calc, arrows.meta, intersections, positioning, patterns, decorations.pathreplacing, backgrounds} %
\usepackage{selectp} %
\usepackage{mathtools}
\usepackage{subcaption}
\usepackage{amsfonts}
\usepackage[noend]{algorithm2e}
\usepackage{calculator}
\usepackage{siunitx}
\usepackage{stfloats}
\usepackage{soul}  %
\usepackage[absolute,overlay]{textpos}
\usepackage{cite}
\usepackage{makecell, multirow}
\usepackage{comment}
\setcellgapes{1.1pt}
\usepackage{booktabs}
\newcommand{\tabitem}{~~\llap{\textbullet}~~}

\newcommand*\rotTwo{\rotatebox{-90}}
\definecolor{TABLE_GOOD}{RGB}{0,0,0}
\definecolor{TABLE_BAD}{RGB}{0,0,0}
\definecolor{PATH_OKAY}{RGB}{0,0,0}
\definecolor{PATH_COLLISION}{RGB}{230,0,0}
\definecolor{PLOT_RED}{RGB}{214,39,40}
\definecolor{PLOT_BLUE}{RGB}{0,0,150}
\definecolor{LINE_COLOR_RED}{RGB}{214,39,40}
\definecolor{LINE_COLOR_BLUE}{RGB}{0,0,150}
\definecolor{LINE_COLOR_GREEN}{RGB}{44,160,44}
\definecolor{UNSAFE_STATE}{RGB}{255,0,0}
\definecolor{SAFE_STATE}{RGB}{0,255,0}

\definecolor{TABLE_ENTRY_ONE}{RGB}{214,39,40}
\definecolor{TABLE_ENTRY_TWO}{RGB}{0,140,0}
\definecolor{TABLE_ENTRY_THREE}{RGB}{0,0,150}
\definecolor{TABLE_ENTRY_FOUR}{RGB}{139,69,13} %

\definecolor{REFERENCE_PATH_GREEN}{RGB}{65,251,61}
\definecolor{GENERATED_PATH_RED}{RGB}{254,24,1}

\definecolor{TABLE_ENTRY_STATE_ACTION_RISK}{RGB}{0,0,150}
\definecolor{TABLE_ENTRY_STATE_RISK}{RGB}{0,140,0}

\RestyleAlgo{ruled}

\tikzstyle{image_frame} = [rounded corners=1.5pt, inner sep=0.25pt, draw=black]

\title{\LARGE \bf
Jerk-limited Traversal of One-dimensional Paths and its Application to Multi-dimensional Path Tracking  
}

\author{Jonas C. Kiemel$^{1}$ and Torsten Kröger %
\thanks{\protect\hypertarget{link:author}{$^{1}$}Institute for Anthropomatics and Robotics – Intelligent Process Automation and Robotics (IAR-IPR), Karlsruhe Institute of Technology (KIT), jonas.kiemel@kit.edu %
}%
}

\begin{document}

\maketitle
\thispagestyle{empty}
\pagestyle{empty}

\begin{textblock*}{14.9cm}(3.2cm,0.75cm) 
	{\footnotesize © 2024 IEEE.  Personal use of this material is permitted.  Permission from IEEE must be obtained for all other uses, in any current or future media, including reprinting/republishing this material for advertising or promotional purposes, creating new collective works, for resale or redistribution to servers or lists, or reuse of any copyrighted component of this work in other works.}
\end{textblock*}

\begin{abstract}

In this paper, we present an iterative method to quickly traverse multi-dimensional paths considering jerk constraints. 
As a first step, we analyze the traversal of each individual path dimension.
We derive a range of feasible target accelerations for each intermediate waypoint of a one-dimensional path using a binary search algorithm. %
Computing a trajectory from waypoint to waypoint leads to the fastest progress on the path when selecting the highest feasible target acceleration. %
Similarly, it is possible to calculate a trajectory that leads to minimum progress along the path.
This insight allows us to control the traversal of a one-dimensional path in such a way that a reference path length of a multi-dimensional path is approximately tracked over time.
In order to improve the tracking accuracy, we propose an iterative scheme to adjust the temporal course of the selected reference path length. 
More precisely, the temporal region causing the largest position deviation is identified and updated at each iteration. %
In our evaluation, we thoroughly analyze the performance of our method using seven-dimensional reference paths with different path characteristics.  %
We show that our method manages to quickly traverse the reference paths and compare the
required traversing time and the resulting path accuracy with other state-of-the-art approaches.

\end{abstract}
\begin{figure}[t]
    \vspace{-0.125cm}
    \begin{tikzpicture}[scale=1.0, every node/.style={scale=1.0}, node distance=2cm]
            \node at (-0.0, -0.1cm)(path_sections) {\includegraphics[trim=0 260 0 0, clip, width=\linewidth]{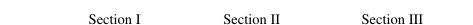}};
            \node[below of=path_sections, node distance=1.2cm](path_position) {\includegraphics[trim=0 153 0 0, clip, width=\linewidth]{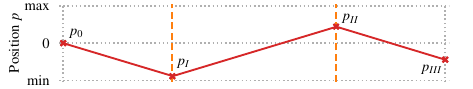}};
            \node[below of=path_position, node distance=1.0cm](path_legend) {\includegraphics[trim=0 0 0 185, clip, width=\linewidth]{figures/header_path/single_dimension_spline.pdf}};
            \node[below of=path_legend, node distance=3.95cm](path_time) {\includegraphics[trim=0 0 0 0, clip, width=\linewidth]{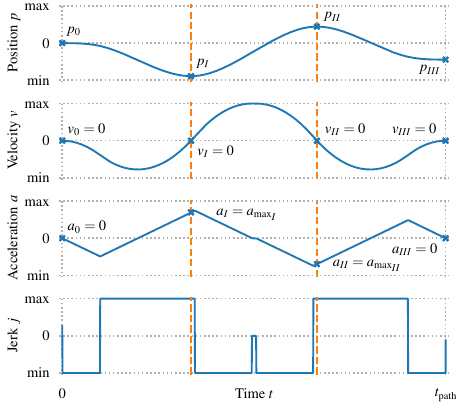}};
            
             \draw [line width=0.25pt] ($(path_position.0)+(-0.2425cm, -0.685cm)$) -- + (0.0cm, 1.4cm) node[pos=0.5, align=center, scale=0.7, fill=white]{\rotTwo{Path\strut}};
             \draw [line width=0.25pt] ($(path_time.45)+(+0.35cm, -7.0cm)$) -- + (0.0cm, 6.8cm) node[pos=0.5, align=center, scale=0.7, fill=white]{\rotTwo{Time-optimized  path parameterization\strut}};
    \end{tikzpicture}
    \vspace{-0.6cm}
	\caption{%
    A time parameterization for an exemplary one-dimensional path with an initial position $p_0$, final position $p_{III}$ and  two intermediate \mbox{positions $p_I$ and $p_{II}$}.}
	\label{fig:header}
	\vspace{-0.5cm}
\end{figure}

\section{INTRODUCTION}
One of the most common tasks in industrial robotics is to follow a given reference path.
The problem of traversing a reference path in a time-optimal manner, while respecting the kinematic limits of the robot joints, is known as time-optimal path parameterization (TOPP).
For velocity and acceleration constraints, the problem has been studied extensively, and open source implementations are readily available \cite{kunz2012time, pham2018Toppra}.
However, constraints on the derivative of the acceleration, commonly known as jerk, are often ignored. 
By considering jerk constraints, wear and tear on the mechanical components can be reduced, which not only increases the reliability of a robotic system but also contributes to its overall cost-effectiveness.
While TOPP with jerk constraints is still an ongoing research topic, progress has been made in addressing a related problem. Specifically, it is possible to compute a time-optimal trajectory considering jerk constraints when provided with an initial kinematic state and a desired target kinematic state of a robot joint \cite{berscheid2021jerk}. These kinematic states consist of a position $p$, velocity $v$ and acceleration $a$. 
With this in mind, we analyze the problem of traversing a one-dimensional path subject to jerk constraints. 
As shown in Fig.~\ref{fig:header}, a one-dimensional path can be described by its initial position $p_0$, its final position $p_{III}$ and intermediate positions ($p_I$ and $p_{II}$) at which the direction of the path changes.
The corresponding velocities $v_0$, $v_{I}$, $v_{II}$, $v_{III}$ need to be zero. The same applies to the initial acceleration $a_0$ and the final acceleration $a_{III}$.
To fully specify kinematic target states for the intermediate points $p_I$ and $p_{II}$, appropriate accelerations  $a_{I}$ and $a_{II}$ need to be found.
The main contributions of this paper can be \mbox{summarized as follows:}
\begin{itemize}
    \item We show how a binary search can be used to compute a range of feasible target accelerations for each intermediate point of a one-dimensional path. Selecting the highest feasible acceleration of each range results in a time-optimized traversal of the reference path (see $a_{\mathrm{max}_{I}}$ and $a_{\mathrm{max}_{II}}$ in Fig.~\ref{fig:header}).
    
    \item Based on these results, we compute trajectories leading to minimum and maximum progress along the one-dimensional path.  
    \item Using these trajectories, we propose an iterative scheme to quickly traverse a multi-dimensional path and benchmark the resulting traversing time and path accuracy with other state-of-the-art methods.
\end{itemize}

\section{Related work} \vspace{-1.0cm}
Early research on time-optimal path parameterization (TOPP) for multi-dimensional paths dates back to the 1980s \cite{bobrow1985time, shin85Time}.
Nowadays, common TOPP algorithms are designed to handle both first-order and second-order constraints, ensuring that the joint velocity and joint acceleration stay within predefined limits. 
An exemplary implementation of such an algorithm based on reachability analysis is provided by the open-source library TOPP-RA~\cite{pham2018Toppra}, which serves as a benchmark for our evaluation.

In order to consider jerk constraints, a TOPP algorithms supporting third-order constraints is required. 
While several partial solutions based on numerical integration have been proposed  \cite{tarkiainen1993time, mattmuller2009calculating, pham2017structure}, an efficient algorithm for the general problem is still the subject of ongoing research. 
Singularities pose a major challenge in this context, as they make numerical integration difficult.

TOPP with first-order constraints and second-order constraints can also be effectively handled by convex optimization~\cite{verscheure2009time}. 
However, the inclusion of third-order constraints leads to a non-convex optimization problem \cite{verscheure2009time, zhang2013practical}. 
To address this issue, a convex approximation based on linear constraints is proposed in~\cite{zhang2013practical}.
As an alternative, third-order constraints can be approximated by adding a penalty term  to the objective function of the optimization problem \cite{gasparetto2008technique}.

While TOPP is typically performed offline, there are also online methods capable of considering jerk constraints. In \cite{lange2015trajectory} and \cite{lange2015path}, a jerk-limited path tracking technique is proposed, however, without focusing on the traversing time. 
As mentioned in the introduction, it is also possible to compute a time-optimal trajectory from one kinematic state to another considering jerk constraints \cite{macfarlane2003jerk, haschke2008line, broquere2008soft, kroger2011opening, hehn2015real, beul2016analytical, berscheid2021jerk}.
The Reflexxes motion library~\cite{kroger2011opening} can process a combination of a target position and a target velocity as a target state.
A more recent development, the Ruckig library \cite{berscheid2021jerk}, goes a step further by also supporting target accelerations.
We use the open-source community version of Ruckig as a backend for our calculations.
There is also a commercial closed-source version called Ruckig pro, which additionally supports intermediate waypoints.
By densely sampling waypoints from a given path, Ruckig pro can be used to generate an approximate path parameterization. %
In our evaluation, we consider the results of this approximation as a benchmark. 

In \cite{kiemel2021learning}, it is shown that an upper and a lower trajectory can be computed for each joint such that position, velocity, acceleration and jerk constraints are not violated.
This insight is used to construct a continuous action space for reinforcement learning in which each action leads to a feasible trajectory. 
Using this action space, it is possible to train a neural network to follow a path without violating jerk constraints \cite{kiemel2022path}. 
As a reference, we also provide the results of this method in our evaluation section.

\begin{figure}[t]
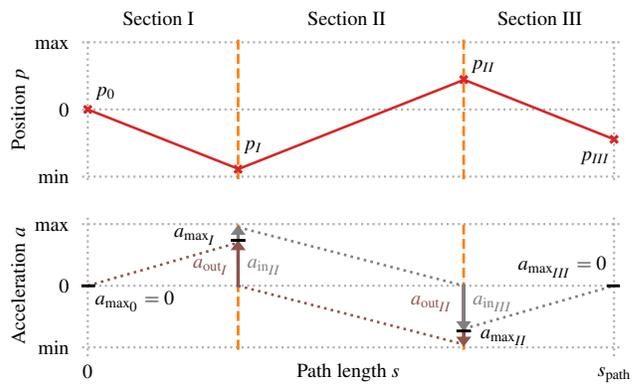

    \vspace{-0.125cm}
    \begin{tikzpicture}[scale=1.0, every node/.style={scale=1.0}, node distance=2cm]
            \node at (-0.0, -0.1cm)(path_position_and_acc) {\includegraphics[trim=0 142 0 0, clip, width=\linewidth]{figures/maximum_accelerations/path_with_acc.pdf}};
            \node[below of=path_position_and_acc, node distance=2.65cm](path_legend) {\includegraphics[trim=0 5 0 256, clip, width=\linewidth]{figures/maximum_accelerations/path_with_acc.pdf}};
    \end{tikzpicture}
    \vspace{-0.7cm}
	\caption{%
	The range of valid target accelerations for each waypoint $p_0$, $p_{I}$, $p_{II}$, $p_{III}$ includes zero as lower limit and $a_{\mathrm{max}_{0}}$, $a_{\mathrm{max}_{I}}$, $a_{\mathrm{max}_{II}}$ and $a_{\mathrm{max}_{III}}$ as upper limit, respectively.} %
	\label{fig:maximum_accelerations}
	\vspace{-0.4cm}
\end{figure}

\section{Approach}
\subsection{Overview}
In a first step, we analyze the traversal of a one-dimensional path subject to the following constraints:
\begin{alignat}{3}
v_{\mathrm{min}} &{}\le{}& \dot{p} &{}\le{}& v_{\mathrm{max}}\phantom{,}    \label{eq:constraint_v}\\
a_{\mathrm{min}} &{}\le{}& \ddot{p}&{}\le{}& a_{\mathrm{max}}\phantom{,}  \label{eq:constraint_a}  \\
j_{\mathrm{min}} &{}\le{}& \dddot{p} &{}\le{}& j_{\mathrm{max}},  \label{eq:constraint_j}
\end{alignat}
where $p$, $v$, $a$ and $j$ stand for position, velocity, acceleration and jerk, respectively.

As shown in Fig. \ref{fig:maximum_accelerations} and explained in (\ref{sec:target_accelerations}), a range of feasible target accelerations can be computed for each intermediate waypoint of a one-dimensional path. Given a feasible target acceleration, the Ruckig library \cite{berscheid2021jerk} can be used to compute a time-optimal trajectory from one waypoint to another without violating the constraints (\ref{eq:constraint_v}) - (\ref{eq:constraint_j}). %

In section (\ref{sec:upper_lower_trajectory}), we explain how the traversal on the path can be controlled by repeatedly computing an upper and a lower trajectory. %

In section (\ref{sec:iterative_scheme}), we finally introduce an iterative scheme to closely follow a multi-dimensional path. %

\subsection{Computing feasible target accelerations}
\label{sec:target_accelerations}
In this section, we analyze feasible kinematic states \mbox{($p$, $v$, $a$)} for each waypoint $p_0$, $p_{I}$, $p_{II}$, $p_{III}$ of a one-dimensional path shown in Fig.~\ref{fig:maximum_accelerations}.
While our exemplary path has two intermediate waypoints $p_{I}$ and $p_{II}$, the same principle can be applied to arbitrary one-dimensional paths. 
It is evident that the velocity and acceleration of the first waypoint $p_0$ and the last waypoint $p_{III}$ must be zero.
The intermediate waypoints $p_{I}$ and $p_{II}$ are local extrema of the path. Consequently, their corresponding velocities also need to be zero.  
In order to compute feasible target accelerations for the intermediate waypoints, we first consider the movement from $p_0$ to $p_{I}$. Since $p_{I}$ is a local minimum of the path, the target acceleration must be greater than or equal to zero. Starting from stillstand, a target acceleration of zero is always possible as it results in a normal point-to-point motion. 
\begin{figure}[t]
    \vspace{-0.325cm}
    \begin{tikzpicture}
	    \definecolor{MATPLOTLIB_DARKGREEN}{RGB}{0, 100, 0}
	      \definecolor{MATPLOTLIB_DARKRED}{RGB}{139, 0, 0}
                \node at ($(0.0cm, 0.0cm)+(0.0cm, -0.1cm)$) {}; %
            	\node[scale=0.75] at ($(0.0cm, 0.0cm)+(-3.3cm, -0.0cm)$){$\;\;\;\;\;\;\;\;\;\;\;$(a) $\boldsymbol{\textcolor{MATPLOTLIB_DARKRED}{a_{\mathrm{in}}}} = a_{\mathrm{max}}$\strut};
                \node[scale=0.75] at ($(0.0cm, 0.0cm)+(-0.15cm, -0.0cm)$){(b) $\boldsymbol{\textcolor{MATPLOTLIB_DARKGREEN}{a_{\mathrm{in}}}} = 0.5 \cdot a_{\mathrm{max}}$\strut};
                \node[scale=0.75] at ($(0.0cm, 0.0cm)+(2.5cm, -0.0cm)$){(c) $\boldsymbol{\textcolor{MATPLOTLIB_DARKGREEN}{a_{\mathrm{in}}}} = 0.75 \cdot a_{\mathrm{max}}$\strut};
    \end{tikzpicture}
    \includegraphics[trim=0 0 0 0, clip, height=0.65\linewidth]{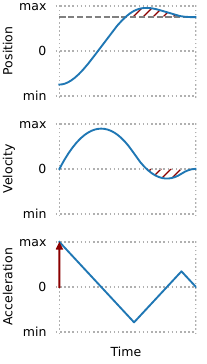}
    \hspace{-0.005\linewidth}
    \includegraphics[trim=8 0 23 0, clip, height=0.65\linewidth]{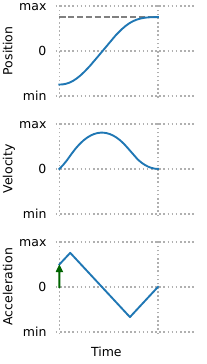}
    \hspace{0.02\linewidth}
    \includegraphics[trim=8 0 23 0, clip, height=0.65\linewidth]{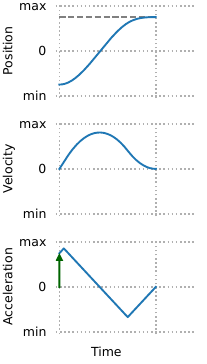}
	\caption{%
	Binary search to find $a_{\mathrm{in}_{\mathrm{max}}}$ assuming $a_{\mathrm{out}} = 0$. 
    After the steps (a), (b) and (c), it can be said that $a_{\mathrm{in}_{\mathrm{max}}}$ is greater than or equal to $0.75 \cdot a_{\mathrm{max}}$ but less than $a_{\mathrm{max}}$.}
	\label{fig:binary_search}
	\vspace{-0.5cm}
\end{figure}
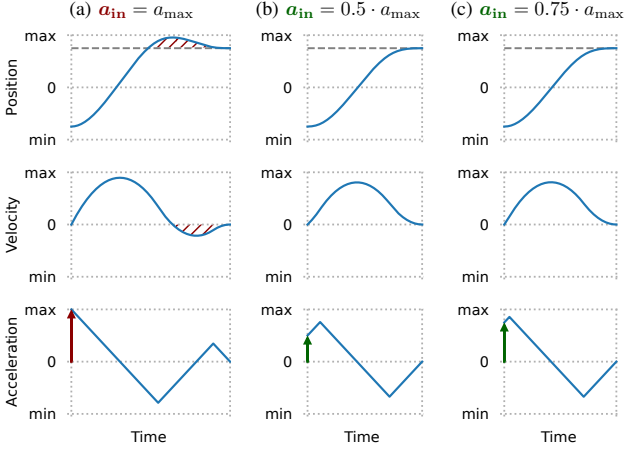
The range of potential target accelerations $a_{\mathrm{out}_{I}}$ is continuous. For that reason, all potential target accelerations are known, once the maximum target acceleration $a_{\mathrm{out}_{\mathrm{max}_{I}}}$ is found. 
However, it could happen that $a_{\mathrm{out}_{\mathrm{max}_{I}}}$ is greater than the maximum input acceleration of the next section $a_{\mathrm{in}_{\mathrm{max}_{II}}}$. 
Consequently, the maximum target acceleration $a_{\mathrm{max}_{I}}$ is calculated as follows:
\begin{align}
    a_{\mathrm{max}_{I}} = \min(a_{\mathrm{out}_{\mathrm{max}_{I}}}, a_{\mathrm{in}_{\mathrm{max}_{II}}})
\end{align}
Both, $a_{\mathrm{out}_{\mathrm{max}_{I}}}$ and $a_{\mathrm{in}_{\mathrm{max}_{II}}}$ are determined by a binary search. 
In Fig. \ref{fig:binary_search}, the principle is illustrated for the maximum input acceleration $a_{\mathrm{in}_{\mathrm{max}}}$. In a first step (a), $a_{\mathrm{in}_{\mathrm{max}}}$ is assumed to be $a_{\mathrm{max}}$. Using Ruckig, a trajectory from the current waypoint $p_\mathrm{current}$ to the next waypoint $p_\mathrm{next}$ is computed, selecting both the target velocity and the target acceleration to be zero. If the resulting trajectory is valid, the maximum input acceleration is $a_{\mathrm{max}}$. If not, another test~(b) is performed, assuming $a_{\mathrm{in}_{\mathrm{max}}} = 0.5 \cdot a_{\mathrm{max}}$. A trajectory is considered as valid if:
\begin{itemize}
    \item The kinematic limits (\ref{eq:constraint_v}) - (\ref{eq:constraint_j}) are not violated. 
    \item The velocity is never negative if $p_\mathrm{next} > p_\mathrm{current}$ or never positive if $p_\mathrm{next} < p_\mathrm{current}$. 
\end{itemize}
In case of (a), the position overshoots the target. 
Consequently, the trajectory is considered as invalid as the velocity must become negative to compensate for the overshoot.
In step (b) and step (c), the resulting trajectory is considered as valid. As a result, $a_{\mathrm{in}_{\mathrm{max}}}$ must be greater than or equal to $0.75 \cdot a_{\mathrm{max}}$ but less than $a_{\mathrm{max}}$.
The binary search is continued until $a_{\mathrm{in}_{\mathrm{max}}}$ is approximated sufficiently well. 
\begin{figure}[t]
    \vspace{-0.26cm}
    \begin{tikzpicture}[scale=1.0, every node/.style={scale=1.0}, node distance=2cm]
            \node at (0, 0.0cm) { \includegraphics[trim=0 2 0 4, clip, width=\linewidth]{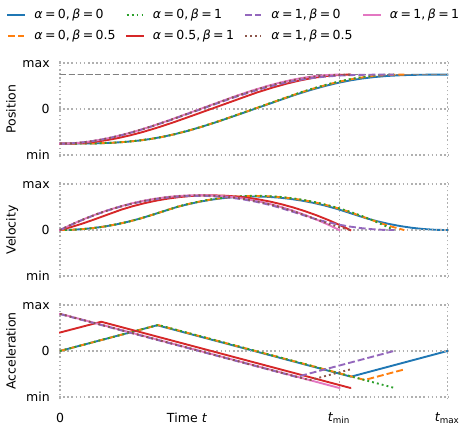}};
             \draw[fill=none, thick](1.885, -3.0675) circle (0.1);
    \end{tikzpicture}
    \vspace{-0.65cm}
	\caption{%
	Resulting trajectories when selecting an input acceleration of $\alpha \cdot a_{\mathrm{in}_{\mathrm{max}}}$ and a target acceleration of $\beta \cdot a_{\mathrm{out}_{\mathrm{max}}}$}
	\label{fig:alpha_beta}
	\vspace{-0.6cm}
\end{figure}

In order to compute $a_{\mathrm{in}_{\mathrm{max}_{II}}}$ for the intermediate section~II, we assume a target acceleration of zero. To compute $a_{\mathrm{out}_{\mathrm{max}_{II}}}$, we assume an input acceleration of zero. 
As shown in Fig. \ref{fig:alpha_beta}, it is nevertheless possible to compute a valid trajectory choosing an input acceleration of $a_{\mathrm{in}_{\mathrm{max}_{II}}}$ and a target acceleration of $a_{\mathrm{out}_{\mathrm{max}_{II}}}$. In fact, any combination of an input acceleration $\alpha \cdot a_{\mathrm{in}_{\mathrm{max}}}$ and a target acceleration $\beta \cdot a_{\mathrm{out}_{\mathrm{max}}}$ leads to a valid trajectory, with $\alpha, \beta \in [0.0, 1.0]$. 
The higher the value of $\alpha$ or $\beta$, the faster the traversal.
The smallest traversal time $t_{\mathrm{min}}$ results from selecting \mbox{$\alpha = \beta = 1.0$}.
Thus, a time-optimized traversal of the path shown in Fig. \ref{fig:header} and Fig. \ref{fig:maximum_accelerations} is achieved by selecting $a_{I} =a_{\mathrm{max}_{I}}$ and $a_{II} =a_{\mathrm{max}_{II}}$.
As highlighted by a circle in Fig. \ref{fig:alpha_beta}, the acceleration of the fastest trajectory, shown in pink, slightly goes up before reaching $a_{\mathrm{out}_{\mathrm{max}}}$. Thus, a slightly higher target acceleration could be selected, leading to a slightly smaller traversal time. However, this target acceleration could no longer be reached from an input acceleration of zero.  
In practice, we found the loss of time caused by including zero as a valid input acceleration and a valid target acceleration to be small.

\subsection{Controlling the traversal of a one-dimensional path} 
\label{sec:upper_lower_trajectory}
\begin{figure}[t]
    \vspace{-0.4cm}
     \begin{tikzpicture}
	    \definecolor{MATPLOTLIB_PURPLE}{RGB}{148, 103, 189}
	      \definecolor{MATPLOTLIB_CYAN}{RGB}{23, 190, 207}
         \node at (-0.0, -0.1cm)(position) {\includegraphics[trim=0 175 0 0, clip, width=\linewidth]{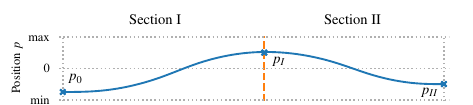}};
        \node[below of=position, node distance=2.2cm](acceleration) {\includegraphics[trim=0 90 0 95, clip, width=\linewidth]{figures/mapping_factor/mapping_factor_0.7.pdf}};
        \node[below of=acceleration, node distance=1.8cm](path_length) {\includegraphics[trim=0 0 0 174, clip, width=\linewidth]{figures/mapping_factor/mapping_factor_0.7.pdf}};
                \node at ($(0.0cm, 0.0cm)+(0.0cm, -0.0cm)$) {}; %
               \draw [draw=MATPLOTLIB_PURPLE, dash pattern=on 3pt off 1.5pt, line width=1.0pt, scale=0.7] ($(position.center)+(-3.0cm, -1.65cm)$) -- + (0.45cm, 0cm) node[pos=1, right, yshift=0.025cm, align=left, scale=0.8]{$a_{\mathrm{lower}_{\Delta t}}$\strut};
              \draw [draw=MATPLOTLIB_CYAN, dash pattern=on 3pt off 1.5pt, line width=1.0pt, scale=0.7] ($(position.center)+(2.075cm, -1.65cm)$) -- + (0.45cm, 0cm) node[pos=1, right, yshift=0.025cm, align=left, scale=0.8]{$a_{\mathrm{upper}_{\Delta t}}$\strut};
               \draw [{Latex[length=0.5mm, width=0.6mm]}-{Latex[length=0.5mm, width=0.5mm]}, draw=black, line width=0.4pt, scale=0.7] ($(acceleration.center)+(-4.55cm, -0.32cm)$) -- + (0.175cm, 0cm);
    \end{tikzpicture}
    \vspace{-0.8cm}
	\caption{%
	Traversing a path with a fixed mapping factor of $0.7$.}
	\label{fig:mapping_factor}
	\vspace{-0.4cm}
\end{figure}

In Fig. \ref{fig:mapping_factor},  a path with one intermediate waypoint $p_{I}$ is traversed. 
As explained before, a time-optimized path traversal
is composed of the following two parts: A trajectory from ($p_{0}$, $0$, $0$) to ($p_{I}$, $0$, $a_{\mathrm{max}_{I}}$), followed by a trajectory from \mbox{($p_{I}$, $0$, $a_{\mathrm{max}_{I}}$)} to ($p_{II}$, $0$, $0$).
We call the resulting trajectory upper trajectory. 
In contrast, the smallest feasible progress on the path is attained through a so-called lower trajectory. 
At the beginning of a section, the lower trajectory corresponds to a simple braking trajectory. 
The braking trajectory can be calculated with Ruckig by specifying a target velocity and a target acceleration of zero. 
However, if the resulting braking trajectory does not stay on the desired path, the calculation is adjusted.
More precisely, when traversing section I, a target state ($p_{I}$, $0$, $a_{\mathrm{min}_{I}}$) is selected.  The acceleration $a_{\mathrm{min}_{I}}$ is less than or equal to $a_{\mathrm{max}_{I}}$ and is computed in a similiar way using a binary search.
As a next step, we discretize the time $t$ into small time steps with a time distance of $\Delta t$ and compute the kinematic states ($p_{\mathrm{lower}_{\Delta t}}$, $v_{\mathrm{lower}_{\Delta t}}$, $a_{\mathrm{lower}_{\Delta t}}$) and ($p_{\mathrm{upper}_{\Delta t}}$, $v_{\mathrm{upper}_{\Delta t}}$, $a_{\mathrm{upper}_{\Delta t}}$) at the following time step.
As shown in Fig. \ref{fig:mapping_factor}, a position and its corresponding section can be uniquely mapped to a path length $s$. We map  $p_{\mathrm{lower}_{\Delta t}}$ and $p_{\mathrm{upper}_{\Delta t}}$ to $s_{\mathrm{lower}_{\Delta t}}$ and $s_{\mathrm{upper}_{\Delta t}}$ and define:
\begin{align}
    s_{\mathrm{desired}_{\Delta t}} = s_{\mathrm{lower}_{\Delta t}} + m \cdot (s_{\mathrm{upper}_{\Delta t}} - s_{\mathrm{lower}_{\Delta t}}),
\end{align}
with $m$ being a mapping factor $\in [0.0, 1.0]$.

For small time distances $\Delta t$, an intermediate trajectory leading to a path length close to $s_{\mathrm{desired}_{\Delta t}}$ can be found.  
Consequently, the mapping factor allows us to control the traversal of the path. 
After each time step, the lower and the upper trajectory are recomputed. %
As shown in Fig. \ref{fig:mapping_factor}, both trajectories are almost identical prior to a section change. 
Thus, the traversal can hardly be influenced during this phase, which motivates the following iterative scheme.

\begin{figure}[t!]
    \vspace{-0.1cm}
    \begin{tikzpicture}[scale=1.0, every node/.style={scale=1.0}, node distance=2cm]
            \definecolor{MATPLOTLIB_BLUE}{RGB}{31, 119, 180}
	          \definecolor{MATPLOTLIB_OLIVE_YELLOW}{RGB}{197, 197, 63}
            \node at (-0.0, -0.1cm)(path_position) {\includegraphics[trim=0 154 0 1, clip, width=\linewidth]{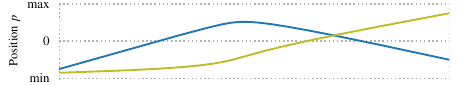}};
            \node[below of=path_position, node distance=1cm](path_legend) {\includegraphics[trim=0 2 0 186, clip, width=\linewidth]{figures/u_delta_tracking/u_delta_tracking_spline.pdf}};
            \node[below of=path_legend, node distance=2.2cm](iteration_1_pos_vel) {\includegraphics[trim=0 285 0 1, clip, width=\linewidth]{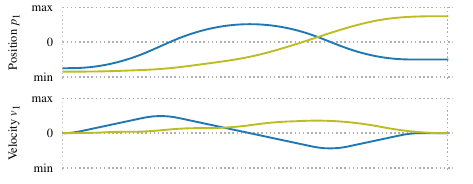}};
             \node[below of=iteration_1_pos_vel, node distance=3.1cm](iteration_1_acc) {\includegraphics[trim=0 240 0 90, clip, width=\linewidth]{figures/u_delta_tracking/u_delta_tracking_iteration_0.pdf}};
            \node[below of=iteration_1_pos_vel, node distance=7.65cm](iteration_1_others) {\includegraphics[trim=0 0 0 175, clip, width=\linewidth]{figures/u_delta_tracking/u_delta_tracking_iteration_0.pdf}};
             \node[below of=iteration_1_acc, node distance=9.15cm](acc_iteration_2) {\includegraphics[trim=0 240 0 90, clip, width=\linewidth]{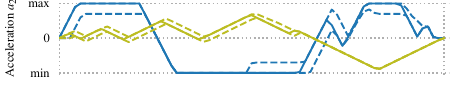}};
             \node[below of=acc_iteration_2, node distance=3.5cm] {\includegraphics[trim=0 8 0 220, clip, width=\linewidth]{figures/u_delta_tracking/u_delta_tracking_iteration_1.pdf}};
             \draw [line width=0.25pt] ($(path_position.0)+(-0.35cm, -0.65cm)$) -- + (0.0cm, 1.35cm) node[pos=0.5, align=center, scale=0.7, fill=white]{\rotTwo{Path\strut}};
             \draw [line width=0.25pt] ($(iteration_1_acc.45)+(3.15cm, -8.075cm)$) -- + (0.0cm, 11.75cm) node[pos=0.5, align=center, scale=0.7, fill=white]{\rotTwo{First iteration\strut}};
            \draw [line width=0.25pt] ($(acc_iteration_2.0)+(-0.35cm, -5.4cm)$) -- + (0.0cm, 6.2cm) node[pos=0.5, align=center, scale=0.7, fill=white]{\rotTwo{Second iteration\strut}};
             \draw[draw=MATPLOTLIB_BLUE, line width=1.5pt, scale=0.7] ($(path_position.center)+(-3.0cm, 1.5cm)$) -- + (0.45cm, 0cm) node[pos=1, right, yshift=0.0cm, xshift=0.03cm, align=left, scale=0.7]{Dimension 1\strut};
            	  \draw [draw=MATPLOTLIB_OLIVE_YELLOW, line width=1.5pt, scale=0.7] ($(path_position.center)+(0.8cm, 1.5cm)$) -- + (0.45cm, 0cm) node[pos=1, right, yshift=0.00cm, xshift=0.03cm, align=left, scale=0.7]{Dimension 2\strut};
            \def\legendXShift{1.8cm}
            \draw [draw=MATPLOTLIB_BLUE, dash pattern=on 3pt off 1.5pt, line width=1.0pt, scale=0.7] ($(iteration_1_acc.center)+(-4.35cm+\legendXShift, +1.74cm)$) -- + (0.4cm, 0cm) node[pos=1, right=0.05cm, yshift=0.045cm, align=left, scale=0.75]{and\strut};
            \draw [draw=MATPLOTLIB_BLUE, dash pattern=on 3pt off 1.5pt, line width=1.0pt, scale=0.7] ($(iteration_1_acc.center)+(-4.35cm+\legendXShift, +1.84cm)$) -- + (0.4cm, 0cm);
            \draw [draw=MATPLOTLIB_OLIVE_YELLOW, dash pattern=on 3pt off 1.5pt, line width=1.0pt, scale=0.7] ($(iteration_1_acc.center)+(-2.95cm+\legendXShift, +1.74cm)$) -- + (0.4cm, 0cm) node[pos=1, right=0.0cm, yshift=0.045cm, align=left, scale=0.75]{: $a_{\mathrm{lower}_{\Delta t}}$ and $a_{\mathrm{upper}_{\Delta t}}$\strut};
            \draw [draw=MATPLOTLIB_OLIVE_YELLOW, dash pattern=on 3pt off 1.5pt, line width=1.0pt, scale=0.7] ($(iteration_1_acc.center)+(-2.95cm+\legendXShift, +1.84cm)$) -- + (0.4cm, 0cm);
    \end{tikzpicture}
    \vspace{-0.40cm}
	\caption{%
	The first two iterations of the proposed iterative tracking scheme shown for an exemplary \mbox{two-dimensional path.}}
	\label{fig:u_delta_tracking}
	\vspace{-0.9cm}
\end{figure}
\subsection{An iterative scheme to track multi-dimensional paths}
\label{sec:iterative_scheme}
In this section, we present an iterative scheme to select the mapping factors such that a multi-dimensional path is approximately tracked. 
Fig. \ref{fig:u_delta_tracking} visualizes the first two iterations of the method for a two-dimensional path. 
As an initial step, we compute the time-optimized traversal of each individual path dimension and identify the slowest dimension. %
For the slowest dimension, we determine the path length $s(t)$ over time corresponding to the time-optimized traversal. %
The path length~$s(t)$ of an individual dimension can be uniquely mapped to a path length~$u(t)$ of the multi-dimensional path. 
The corresponding $u(t)$ of the slowest dimension serves as a reference $u_{\mathrm{ref}_1}$ for the first iteration of our method. 
The basic idea of the next step is to track this reference with the other path dimensions by selecting suitable mapping factors. 
To do so, we define a small range around $u_{\mathrm{ref}}$ where the other dimensions should stay in. The deviation from the reference $u_{\mathrm{ref}}$ is denoted as $\Delta u_{\mathrm{ref}}$. Likewise, the corresponding position deviation is denoted as $\Delta p_{\mathrm{ref}}$. Looking at $\Delta u_{\mathrm{ref}_1}$ in Fig. \ref{fig:u_delta_tracking}, it can be seen that the mapping factors $m_1$ are selected such that dimension~2 oscillates within the desired $\Delta u_{\mathrm{ref}}$ range. However, if it is not possible to keep the dimension within the desired range, the 
mapping factors are selected such that the lower bound of the range is undershot. The corresponding areas and their resulting position deviation are hatched in red. 
We now look for the area outside the desired $\Delta u_{\mathrm{ref}}$ range that causes the largest integrated position deviation of all dimensions. 
As indicated by a dashed black line in Fig.~\ref{fig:u_delta_tracking}, we update the reference $u_{\mathrm{ref}_1}$ in the selected area such that dimension~2 stays within the desired $\Delta u_{\mathrm{ref}}$ range in the following iteration. %

As a consequence, the resulting position deviation after the second iteration $\Delta p_{\mathrm{ref}_2}$ is significantly reduced compared to $\Delta p_{\mathrm{ref}_1}$.
By repeating this iterative procedure, the tracking accuracy can be further increased.
We note that the updated reference can decrease the tracking accuracy of other path dimensions. In practice, we repeat the procedure for a fixed number of iterations and choose the iteration with the best overall tracking performance as the final path parameterization. 

\section{Evaluation}
For our evaluation, we use a KUKA iiwa robot with seven degrees of freedom. 
The selected kinematic limits for each joint are shown in TABLE \ref{table:kinematic_limits}. If not noted otherwise, the jerk limits from the table apply. However, for a more thorough evaluation, we additionally perform experiments with different jerk limits. In these cases, we report a so-called jerk limit factor that is multiplied with the jerk limits given in TABLE \ref{table:kinematic_limits}.

\begin{table}[h]
    \vspace{-0.1cm}
     \caption{Kinematic limits of the robot joints.}
     \vspace{-0.15cm}
    \makegapedcells
\begin{tabular*}{0.49\textwidth}{@{}p{25mm}p{4.5mm}p{4.5mm}p{4.5mm}p{4.5mm}p{4.5mm}p{4.5mm}p{4.5mm}} 
    \toprule
\hspace{0.02cm} &\multicolumn{7}{c}{Joint} \\
 &\multicolumn{1}{c}{1}& \multicolumn{1}{c}{2} & \multicolumn{1}{c}{3} & \multicolumn{1}{c}{4} & \multicolumn{1}{c}{5} & \multicolumn{1}{c}{6} & \multicolumn{1}{c}{7}\\
    \hline
\hspace{0.0002cm} Velocity [rad/s]&1.71&1.71&1.74&2.27&2.44&3.14&3.14\\
\hspace{0.0002cm} Acceleration [rad/$\mathrm{s^2}$]&15.0&\hfil 7.5&10.0&12.5&15.0&20.0&20.0\\
\hspace{0.0002cm} Jerk [rad/$\mathrm{s^3}$]&\hfil 300&\hfil 150&\hfil 200&\hfil 250&\hfil 300&\hfil 400&\hfil 400\\
    \bottomrule
    \end{tabular*}
    \vspace{0ex}
\label{table:kinematic_limits}
\vspace{-0.1cm}
\end{table}

\subsection{Time-optimized traversal of one-dimensional paths}
As a first step, we evaluate the traversal of one-dimensional paths. 
For each of the seven robot joints, we generate 200 one-dimensional paths with randomly selected waypoints. To minimize the trajectory duration, we select the maximum acceleration for each intermediate waypoint as derived in section (\ref{sec:target_accelerations}). As a benchmark, we report the results of the closed-source library Ruckig pro.
TABLE \ref{table:single_dof} shows the resulting average trajectory duration and tracking accuracy for several jerk limit factors. As expected, the trajectory duration decreases if the jerk limits are increased. 
However, the relative impact on the trajectory duration decreases for higher jerk limit factors. 
Compared to Ruckig pro, the results of our method are almost identical.

\begin{table}[h]
     \caption{Time-optimized traversal of one-dimensional paths for different jerk limits (\textcolor{TABLE_ENTRY_THREE}{Ours} / \textcolor{TABLE_ENTRY_FOUR}{Ruckig pro}).}
     \vspace{-0.15cm}
    \makegapedcells
\begin{tabular*}{0.49\textwidth}{@{}p{17mm}p{17mm}p{19mm}p{19mm}} 
    \toprule
\hspace{0.02cm} Jerk limit &\multicolumn{1}{c}{Trajectory}& \multicolumn{2}{c}{Path deviation [rad]}\\
\hspace{0.02cm} factor &\multicolumn{1}{c}{duration [s]}& \hfil mean \hfil &  \hfil max \hfil\\
    \hline
\hspace{0.0002cm}\tabitem Factor 1&\hfil \textcolor{TABLE_ENTRY_THREE}{$5.15$} / \textcolor{TABLE_ENTRY_FOUR}{$5.16$} &\textcolor{TABLE_ENTRY_THREE}{$0.0007$} / \textcolor{TABLE_ENTRY_FOUR}{$0.0026$}&\textcolor{TABLE_ENTRY_THREE}{$0.0014$} / \textcolor{TABLE_ENTRY_FOUR}{$0.0052$}\\
\hspace{0.0002cm}\tabitem Factor 2&\hfil \textcolor{TABLE_ENTRY_THREE}{$5.03$} / \textcolor{TABLE_ENTRY_FOUR}{$5.03$} &\textcolor{TABLE_ENTRY_THREE}{$0.0007$} / \textcolor{TABLE_ENTRY_FOUR}{$0.0006$}&\textcolor{TABLE_ENTRY_THREE}{$0.0014$} / \textcolor{TABLE_ENTRY_FOUR}{$0.0012$}\\
\hspace{0.0002cm}\tabitem Factor 3&\hfil \textcolor{TABLE_ENTRY_THREE}{$5.00$} / \textcolor{TABLE_ENTRY_FOUR}{$5.00$} &\textcolor{TABLE_ENTRY_THREE}{$0.0007$} / \textcolor{TABLE_ENTRY_FOUR}{$0.0006$}&\textcolor{TABLE_ENTRY_THREE}{$0.0014$} / \textcolor{TABLE_ENTRY_FOUR}{$0.0012$}\\
\hspace{0.0002cm}\tabitem Factor 40&\hfil \textcolor{TABLE_ENTRY_THREE}{$4.96$} / \textcolor{TABLE_ENTRY_FOUR}{$4.96$}&\textcolor{TABLE_ENTRY_THREE}{$0.0007$} / \textcolor{TABLE_ENTRY_FOUR}{$0.0007$}&\textcolor{TABLE_ENTRY_THREE}{$0.0014$} / \textcolor{TABLE_ENTRY_FOUR}{$0.0013$}\\
    \bottomrule
    \end{tabular*}
    \vspace{0ex}
\label{table:single_dof}
\vspace{-0.375cm}
\end{table}
\begin{table*}[bp!]
    \vspace{-0.2cm}
    \caption{Benchmarking using geometric shapes (Ruckig pro A: 0.1 rad, Ruckig pro B: 0.2 rad).}
    \vspace{-0.15cm}
    \makegapedcells
\begin{tabular*}{\textwidth}{@{}p{16.0mm}p{8mm}p{11mm}p{17mm}p{17mm}p{18mm}p{19.0mm}p{18mm}p{18mm}} 
    \toprule

\multirow[t]{3}{*}{\hspace{0.02cm}}
    & \multicolumn{4}{c}{Trajectory duration [s]} &  \multicolumn{4}{c}{Path deviation (mean / max) [rad]} \\
     & \multicolumn{1}{c}{Ours}  & \multicolumn{1}{c}{TOPP-RA} & \multicolumn{1}{c}{Ruckig pro A}  & \multicolumn{1}{c|}{Ruckig pro B} &  \multicolumn{1}{c}{Ours}  & \multicolumn{1}{c}{TOPP-RA} & \multicolumn{1}{c}{Ruckig pro A}  & \multicolumn{1}{c}{Ruckig pro B} \\
    \hline
\hspace{0.02cm} \tabitem Circle$\!\!\!\!$
    & \hfil $2.84$ & \hfil \textcolor{TABLE_GOOD}{$2.79$} &  \hfil \textcolor{TABLE_BAD}{$3.64$} &  \hfil \textcolor{TABLE_BAD}{$2.91$} &$\!\!0.0041$ / $0.0083$&$\!\!\!\textcolor{TABLE_GOOD}{<\! 10^{-7}}$ / $\textcolor{TABLE_GOOD}{<\! 10^{-6}}$&$\!\!\textcolor{TABLE_GOOD}{0.0037}$ / $\textcolor{TABLE_BAD}{0.0096}$&$\!\!\textcolor{TABLE_BAD}{0.0110}$ / $\textcolor{TABLE_BAD}{0.0229}$\\
\hspace{0.02cm} \tabitem Lemniscate$\!\!\!\!$
    & \hfil $2.83$ & \hfil \textcolor{TABLE_GOOD}{$2.75$} &  \hfil \textcolor{TABLE_BAD}{$3.68$} &  \hfil \textcolor{TABLE_BAD}{$2.95$} &$\!\!0.0045$ / $0.0108$&$\!\!\!\textcolor{TABLE_GOOD}{<\! 10^{-7}}$ / $\textcolor{TABLE_GOOD}{<\! 10^{-6}}$&$\!\!\textcolor{TABLE_BAD}{0.0046}$ / $\textcolor{TABLE_GOOD}{0.0104}$&$\!\!\textcolor{TABLE_BAD}{0.0115}$ / $\textcolor{TABLE_BAD}{0.0235}$\\
\hspace{0.02cm} \tabitem Heart$\!\!\!\!$
    & \hfil $2.78$ & \hfil \textcolor{TABLE_GOOD}{$2.67$} &  \hfil \textcolor{TABLE_BAD}{$3.28$} &  \hfil \textcolor{TABLE_GOOD}{$2.76$} &$\!\!0.0042$ / $0.0147$&$\!\!\textcolor{TABLE_GOOD}{0.0003}$ / $\textcolor{TABLE_GOOD}{0.0008}$&$\!\!\textcolor{TABLE_BAD}{0.0231}$ / $\textcolor{TABLE_BAD}{0.0420}$&$\!\!\textcolor{TABLE_BAD}{0.0080}$ / $\textcolor{TABLE_BAD}{0.0282}$\\
\hspace{0.02cm} \tabitem Triangle$\!\!\!\!$
   & \hfil $1.64$ & \hfil \textcolor{TABLE_GOOD}{$1.24$} &  \hfil \textcolor{TABLE_GOOD}{$1.52$} &  \hfil \textcolor{TABLE_GOOD}{$1.29$} &$\!\!0.0033$ / $0.0058$&$\!\!\!\textcolor{TABLE_GOOD}{<\! 10^{-7}}$ / $\textcolor{TABLE_GOOD}{<\! 10^{-6}}$&$\!\!\textcolor{TABLE_BAD}{0.0038}$ / $\textcolor{TABLE_BAD}{0.0068}$&$\!\!\textcolor{TABLE_BAD}{0.0059}$ / $\textcolor{TABLE_BAD}{0.0228}$\\
    \bottomrule
    \end{tabular*}
\label{table:evaluation_shapes}
\vspace{-0.2cm}
\end{table*}
\begin{figure}[t]
\captionsetup[subfigure]{margin=80pt}
    \vspace{0.1cm}
    \begin{tikzpicture}
                \hspace{0.62cm}
                 \node[text width=4cm] at (0.2cm, 8.35cm) (origin){};
                 \node at ($(origin.center)+(0.0cm, -0.25cm)$) {}; %
                 \draw [draw=REFERENCE_PATH_GREEN, line width=1.5pt, scale=0.7] ($(origin.center)+(-9.3cm, -0.0cm)$) -- + (0.45cm, 0cm) node[pos=1, right, yshift=0.01cm, align=left, scale=0.85]{Reference path};
        
            	    \draw [draw=GENERATED_PATH_RED, line width=1.5pt, scale=0.7] ($(origin.center)+(-5.5cm, -0.0cm)$) -- + (0.45cm, 0cm) node[pos=1, right, yshift=0.00cm, align=left, scale=0.85]{Path generated with our method};
                \vspace*{2cm}
    \end{tikzpicture}
    \hspace*{0.045\textwidth}
	    \begin{subfigure}[c]{0.2\textwidth}
	   \vspace{0.0cm} 
	   \includegraphics[trim=1085 425 1085 575, clip, height=0.9\textwidth]{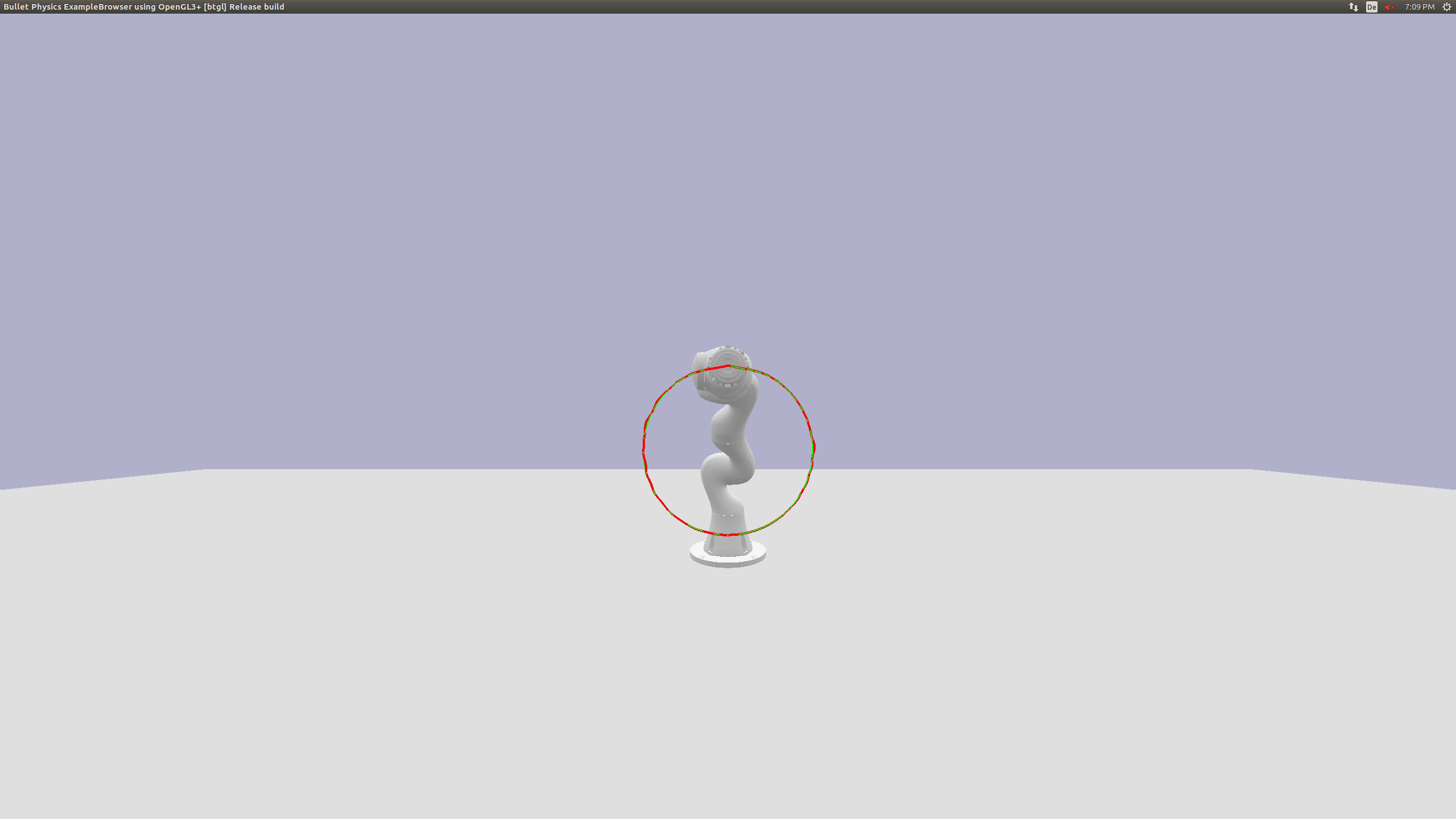}
	   
	   \vspace{-0.425cm}\hspace*{-2.6cm}\subcaptionbox{Circle}[8cm]

	\end{subfigure}
	\begin{subfigure}[c]{0.2\textwidth}
	    \vspace{0.0cm}
	    \includegraphics[trim=1085 425 1085 575, clip, height=0.9\textwidth]{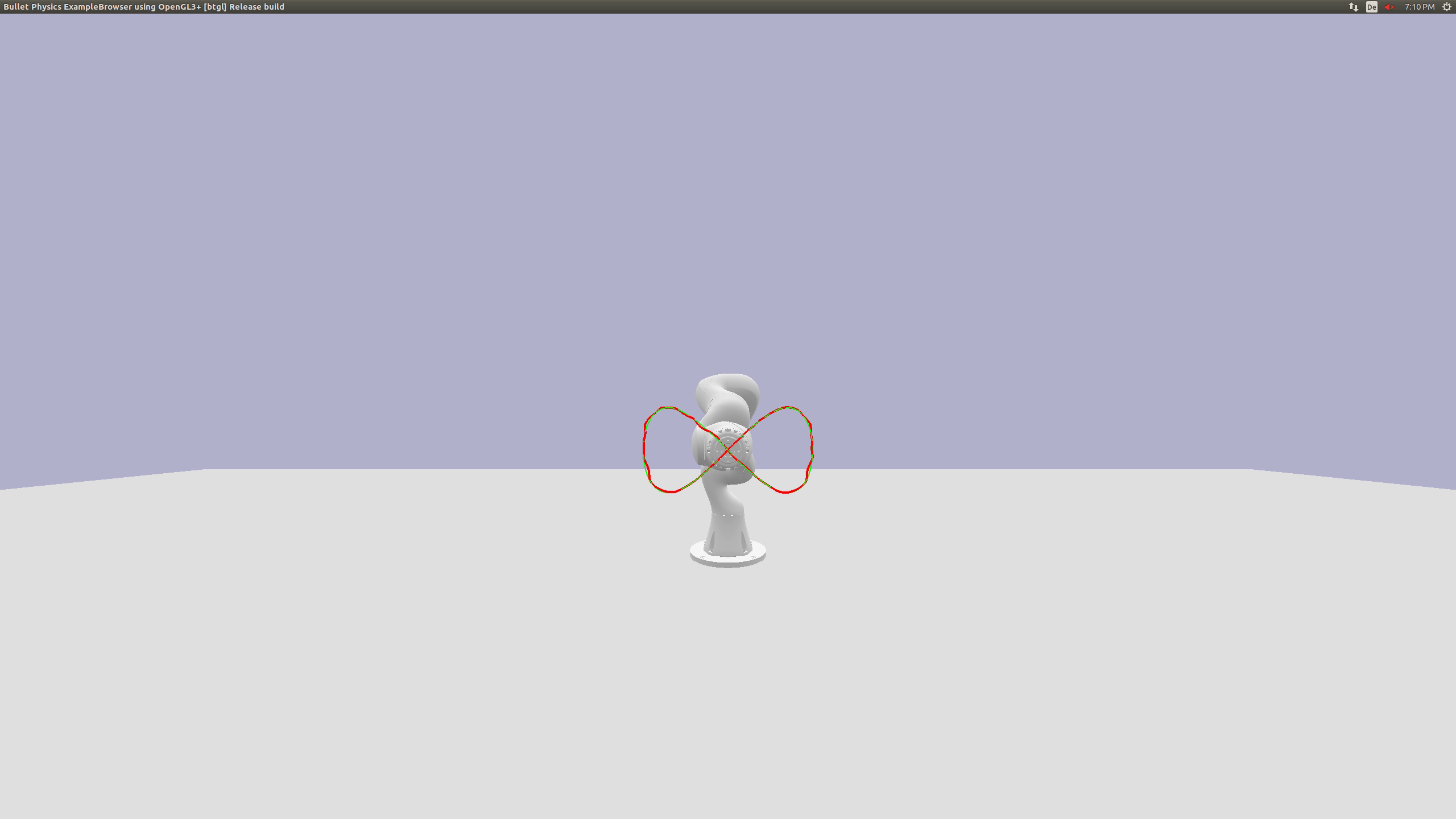}
	    
		\vspace{-0.425cm}\hspace*{-3.1cm}\subcaptionbox{Lemniscate}[9cm]

	\end{subfigure} \vspace{0.1cm}
 
    \hspace*{0.045\textwidth}
    \begin{subfigure}[c]{0.2\textwidth}
	    \vspace{0.0cm}
	    \includegraphics[trim=1085 425 1085 575, clip, height=0.9\textwidth]{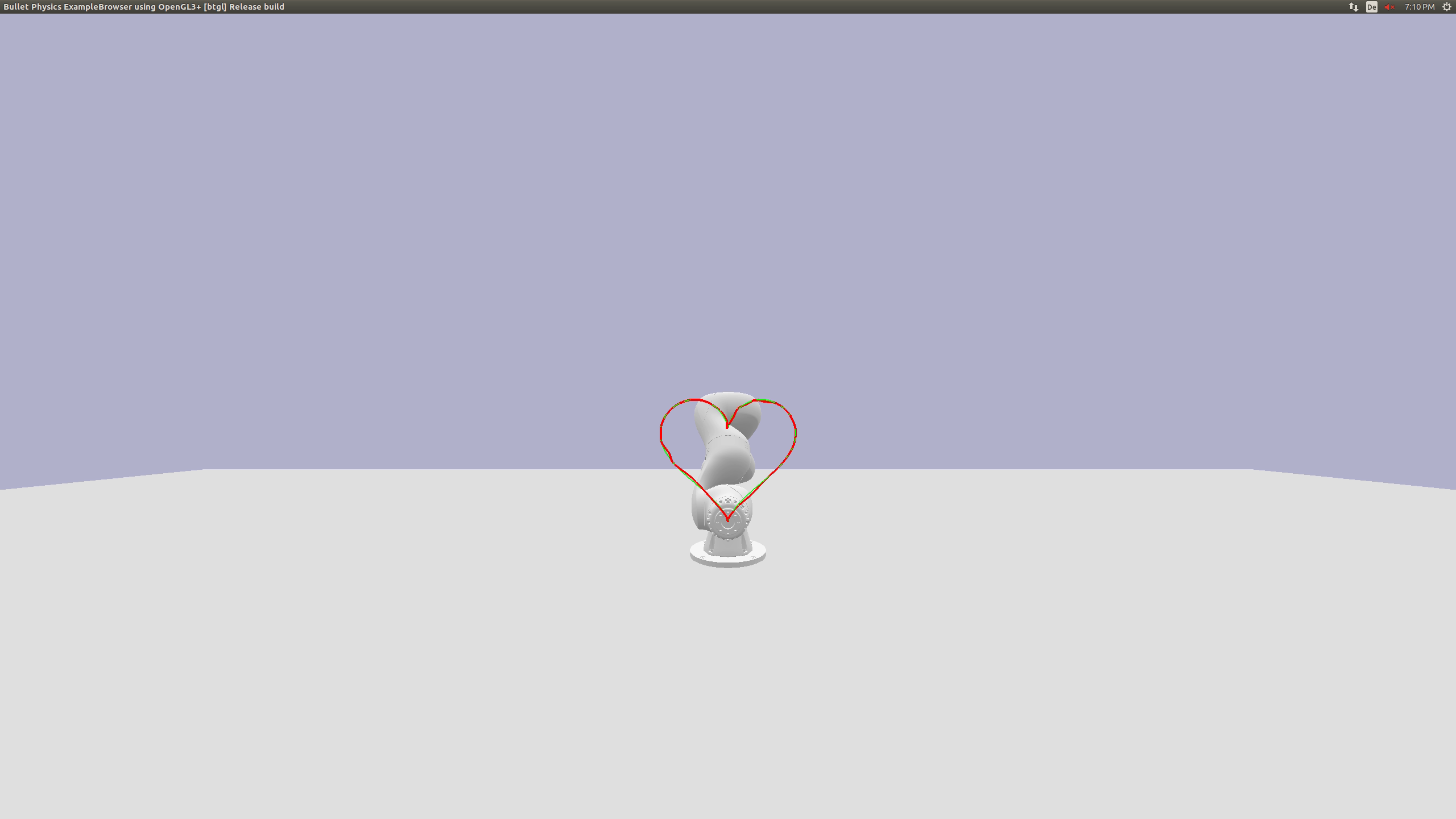}
	    
		\vspace{-0.425cm}\hspace*{-2.63cm}\subcaptionbox{Heart}[8cm]

	\end{subfigure} 
    \begin{subfigure}[c]{0.2\textwidth}
	    \vspace{0.0cm}
	    \includegraphics[trim=1085 425 1085 575, clip, height=0.9\textwidth]{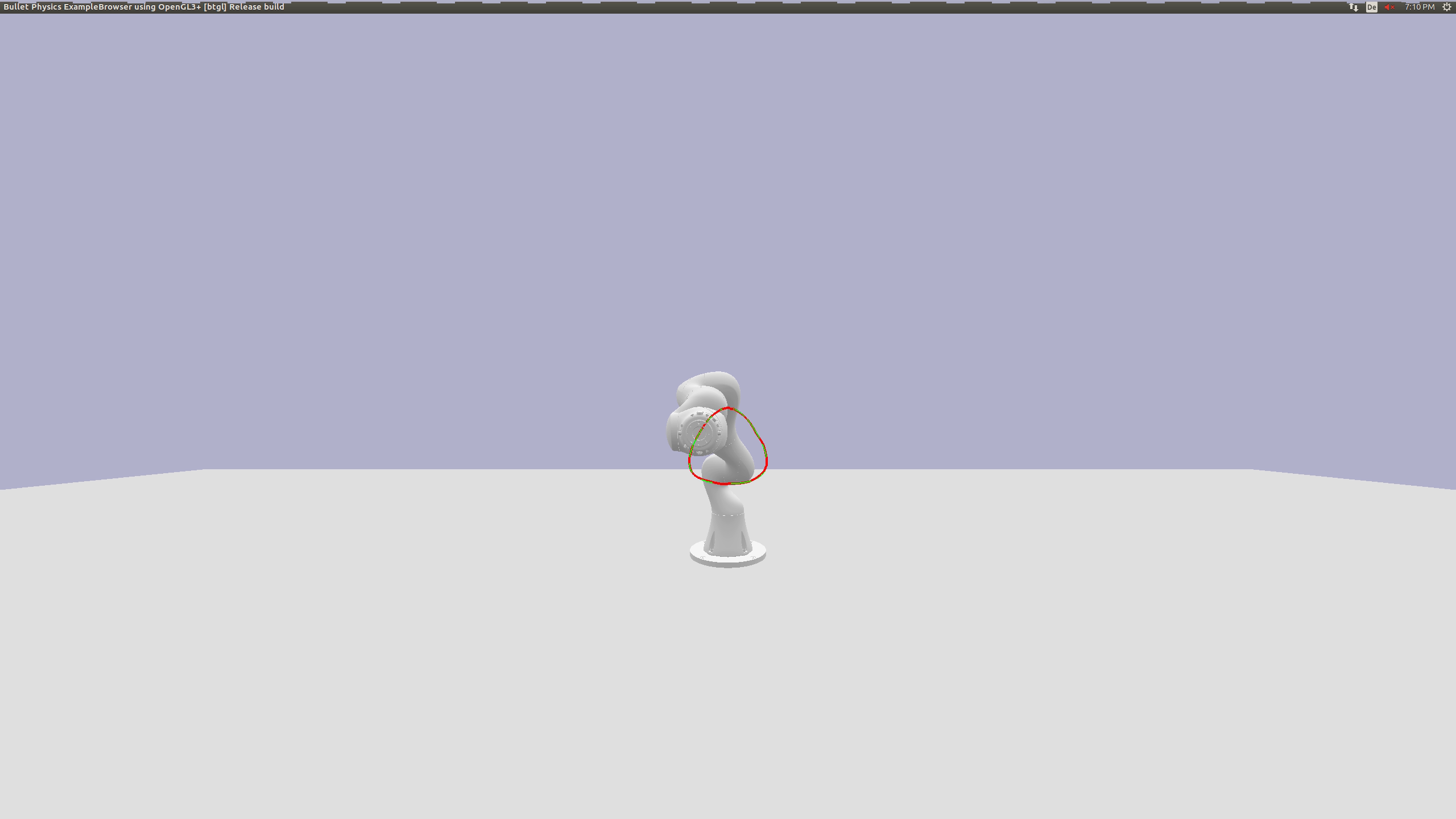}
	    
		\vspace{-0.425cm}\hspace*{-2.65cm}\subcaptionbox{Triangle}[8cm]

	\end{subfigure} 

	 \vspace{0.05cm}   
	\caption{%
Geometric shapes used for our evaluation.}%
	\label{fig:evaluation_shapes}
  \vspace{-0.4cm}
\end{figure}
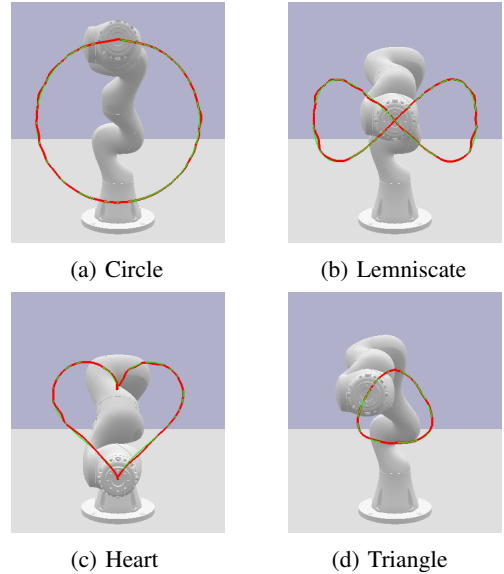
\subsection{Tracking of geometric shapes}
As shown in Fig. \ref{fig:evaluation_shapes}, we apply our iterative tracking method to seven-dimensional paths that resemble geometric shapes in Cartesian space. It can be seen that the resulting Cartesian paths shown in red stay close to the reference paths shown in green. For further visualization, we refer to our accompanying video. 
A quantitative evaluation can be found in TABLE~\ref{table:evaluation_shapes}. 
Using $\Delta t = $ \SI{2.5}{\milli\second} as time step, 20 iterations were performed as described in section \ref{sec:iterative_scheme}.
Finally, the path with the lowest average deviation from the reference path was selected. 
The path deviation is computed by densely sampling waypoints along the reference path and the generated path. 
Next, waypoints sampled at the same path length $u$ are compared by calculating the Euclidean distance between them.
TABLE~\ref{table:evaluation_shapes} shows the mean and the maximum distance calculated in this way.  

We also provide results obtained with TOPP-RA~\cite{pham2018Toppra} and Ruckig pro. With TOPP-RA, the path tracking is very accurate but the jerk limits are ignored. 
Ruckig pro supports jerk limits but is designed to accurately track intermediate waypoints rather than a full path. 
As a reference, we provide the results for an equidistant sampling of waypoints from the reference path using a distance of 0.1 rad~(A) and 0.2 rad~(B).
Overall, our method managed to generate fast trajectories for all of the geometric shapes shown in Fig.~\ref{fig:evaluation_shapes}.

\begin{figure}[t]
\captionsetup[subfigure]{margin=0pt}
    \vspace{0.2cm}
    \hspace{-0.0025\textwidth}
    \begin{subfigure}[c]{0.155\textwidth}
	   \vspace{-0.0cm}
	   \begin{tikzpicture}[scale=1.0, every node/.style={scale=1.0}, node distance=2cm]
            \node[image_frame] at (0, 0.0cm) { \includegraphics[trim=1080 480 1030 470, clip, height=1.025\textwidth]{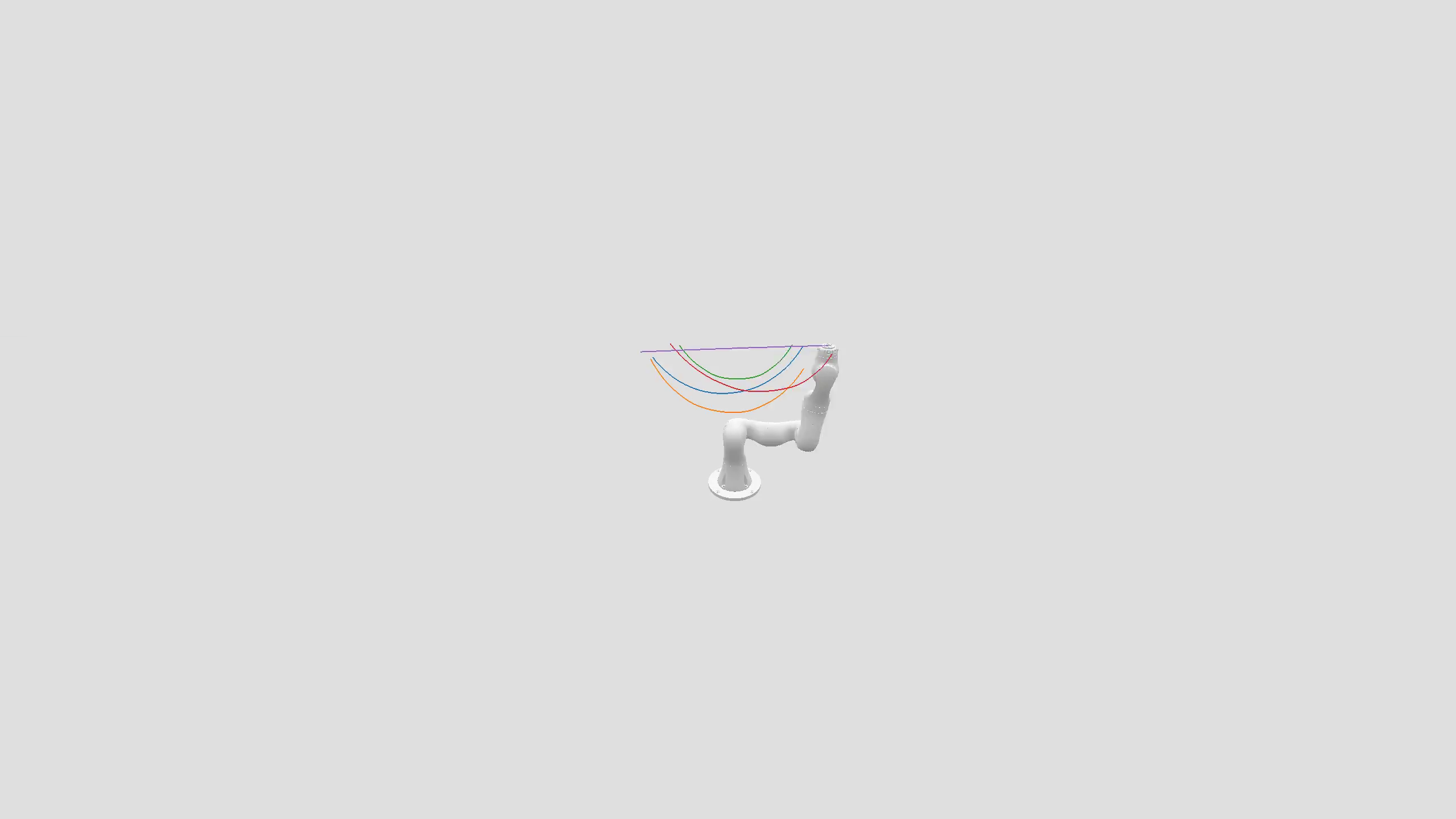}};
        \end{tikzpicture}
	  
	   \vspace{-0.40cm}\hspace*{-1.2cm}\subcaptionbox{Dataset A}[5cm]

	\end{subfigure}
	\begin{subfigure}[c]{0.155\textwidth}
	    \vspace{-0.0cm}
	    \begin{tikzpicture}[scale=1.0, every node/.style={scale=1.0}, node distance=2cm]
            \node[image_frame] at (0, 0.0cm) { \includegraphics[trim=950 340 950 390, clip, height=1.025\textwidth]{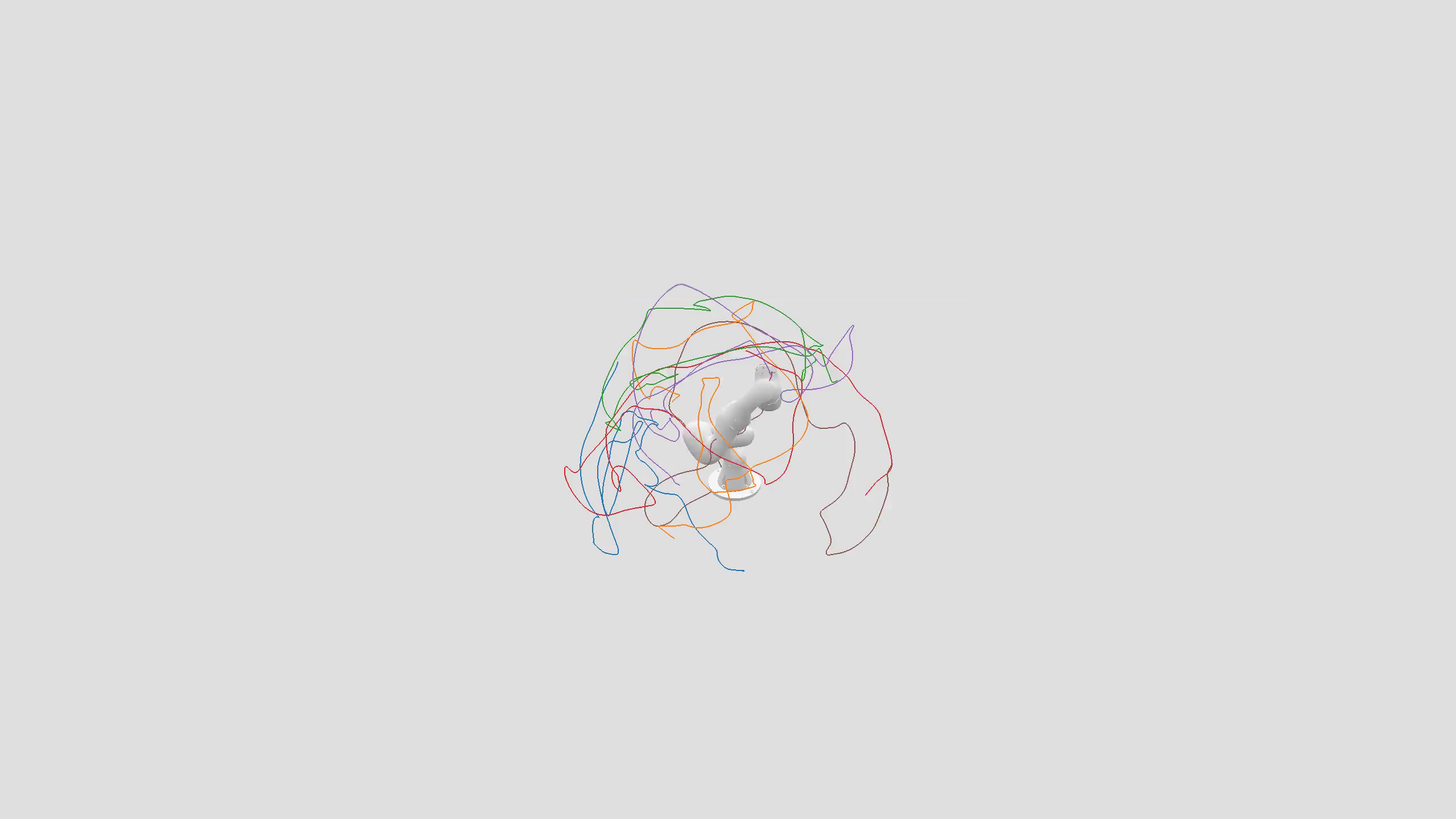}};
        \end{tikzpicture}

		\vspace{-0.4cm}\hspace*{-1.2cm}\subcaptionbox{Dataset B}[5cm]

	\end{subfigure} 
    \begin{subfigure}[c]{0.155\textwidth}
	   \vspace{-0.0cm}
	   \begin{tikzpicture}[scale=1.0, every node/.style={scale=1.0}, node distance=2cm]
            \node[image_frame] at (0, 0.0cm) { \includegraphics[trim=970 440 930 290, clip, height=1.025\textwidth]{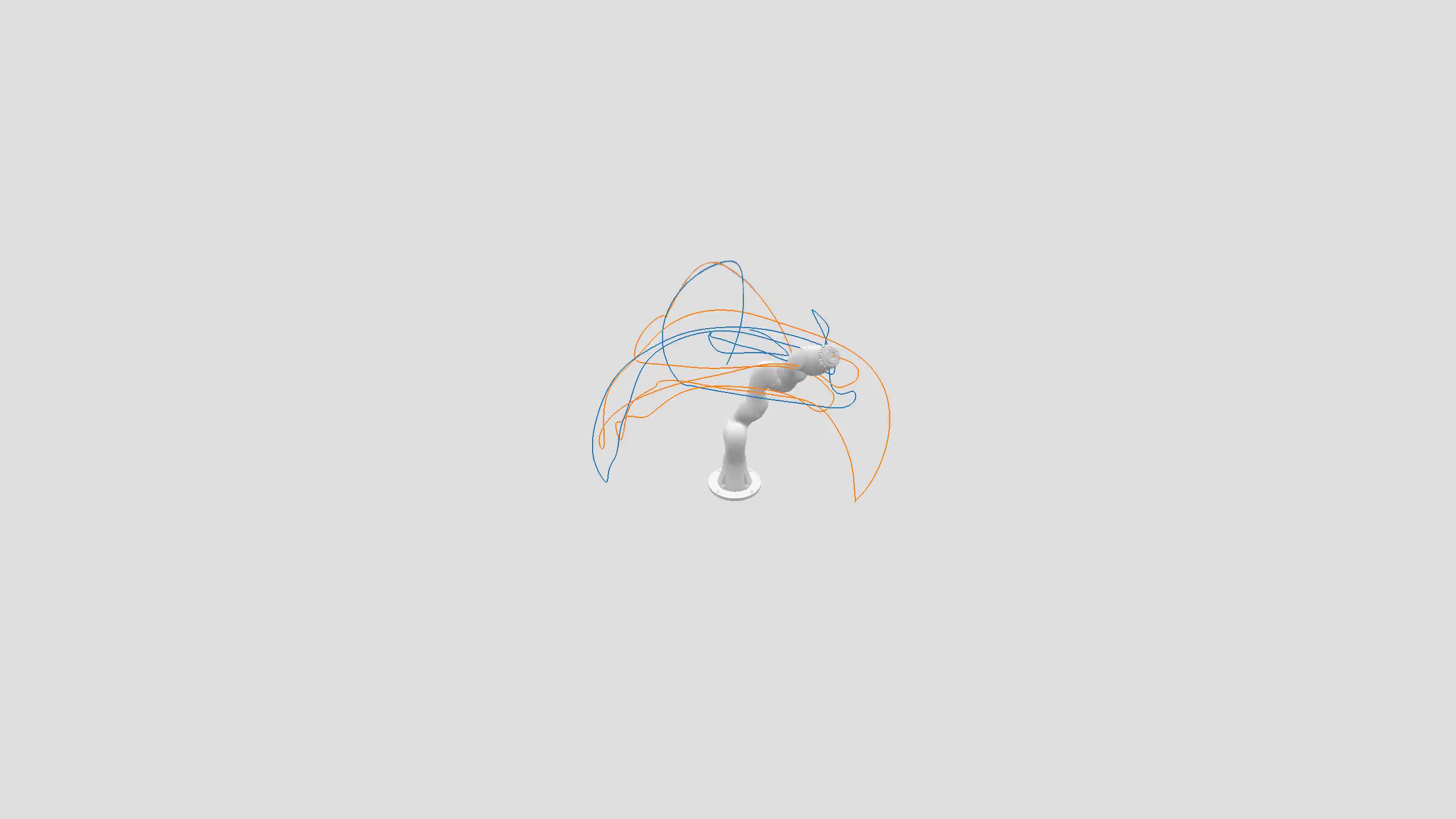}};
        \end{tikzpicture}

	   \vspace{-0.4cm}\hspace*{-1.2cm}\subcaptionbox{Dataset C}[5cm]

	\end{subfigure}

	\caption{Exemplary paths from our datasets.}%
	\label{fig:datasets}
	\vspace{-0.28cm}
\end{figure}

\begin{table*}[t]
    \vspace{-0.2cm}
    \caption{Benchmarking with a method using neural networks based on three datasets with different path characteristics.}
    \vspace{-0.15cm}
    \makegapedcells
\begin{tabular*}{\textwidth}{@{}p{17.0mm}p{18mm}p{18mm}p{13mm}p{20mm}p{22.0mm}p{22.0mm}p{14.5mm}} 
    \toprule

\multirow[t]{3}{*}{\hspace{0.02cm}}
    & \multicolumn{1}{c}{Total path length} & \multicolumn{3}{c}{Trajectory duration [s]} &  \multicolumn{2}{c}{Path deviation (mean / max) [rad]} & \multicolumn{1}{c}{Iteration}\\
     & \multicolumn{1}{c|}{$u_{\mathrm{path}}$ [rad]} & \multicolumn{1}{c}{Slowest dimension}  & \multicolumn{1}{c}{Ours} & \multicolumn{1}{c|}{Neural network}  & \multicolumn{1}{c}{Ours} &  \multicolumn{1}{c|}{Neural network} & \multicolumn{1}{c}{Ours} \\
    \hline
\hspace{0.02cm} \tabitem Dataset A$\!\!\!\!$ & \hfil $3.7$ & \hfil $1.79$ & \hfil $2.06$ &  \hfil \textcolor{TABLE_BAD}{$2.11$} & \hfil  $0.0027$ / $0.0098$& \hfil $\textcolor{TABLE_BAD}{0.0394}$ / $\textcolor{TABLE_BAD}{0.0831}$ & \hfil $9.5$\\
\hspace{0.02cm} \tabitem Dataset B$\!\!\!\!$ & \hfil $22.9$ & \hfil $5.55$ & \hfil $6.69$ &  \hfil \textcolor{TABLE_BAD}{$8.62$} & \hfil  $0.1122$ / $0.2623$& \hfil $\textcolor{TABLE_GOOD}{0.1088}$ / $\textcolor{TABLE_GOOD}{0.2113}$ & \hfil $12.3$\\
\hspace{0.02cm} \tabitem Dataset C$\!\!\!\!$ & \hfil $25.6$ & \hfil $6.68$ & \hfil $7.67$ &  \hfil \textcolor{TABLE_BAD}{$10.51$} & \hfil  $0.0962$ / $0.2283$& \hfil $\textcolor{TABLE_BAD}{0.1227}$ / $\textcolor{TABLE_BAD}{0.2685}$ & \hfil $12.1$\\
    \bottomrule
    \end{tabular*}
\label{table:evaluation_dataset_ours}
\vspace{-0.2cm}
\end{table*}

\subsection{Tracking of paths with different characteristics}
In the following, we evaluate our method using three datasets with different path characteristics. Exemplary paths for each dataset are shown in Fig. \ref{fig:datasets}.
Dataset A contains a wide range of semicircles and straight lines. The paths in dataset B are generated by selecting random joint accelerations. Dataset C is composed of paths obtained by moving between randomly sampled Cartesian target points. 
In \mbox{TABLE \ref{table:evaluation_dataset_ours}}, we report the results for each dataset and compare them with \cite{kiemel2022path}, a method that uses neural networks trained via reinforcement learning to track the reference paths. 
The indicated trajectory duration of the slowest individual path dimension serves as a lower limit for the resulting trajectory duration when considering the multi-dimensional path.
It can be seen, that the presented method generates faster trajectories than \cite{kiemel2022path} for all datasets. In addition, the tracking accuracy for dataset A and dataset C is higher. 
In TABLE \ref{table:evaluation_dataset_ours}, we also specify the average iteration chosen for our evaluation.   

TABLE \ref{table:evaluation_dataset_benchmarking} shows additional results obtained with \mbox{TOPP-RA} and Ruckig pro. 
For dataset A, \mbox{TOPP-RA} generates the fastest trajectories with the highest path accuracy, however, without considering jerk limits. Using Ruckig pro with a waypoint distance of 0.1 rad leads to slower trajectories, but a higher tracking accuracy compared to our method. In contrast, selecting a waypoint distance of 0.15 rad or 0.2 rad leads to faster trajectories but a less accurate path tracking. 
For dataset B and dataset C, the resulting trajectories with TOPP-RA are slower than with our method but the tracking accuracy is higher. %
\begin{table}[h!]
    \vspace{-0.0cm}
     \caption{Additional benchmarking using datasets with different path characteristics}
     \vspace{-0.15cm}
    \makegapedcells
\begin{tabular*}{0.49\textwidth}{@{}p{27mm}p{17mm}p{15mm}p{15mm}} 
    \toprule
\hspace{0.02cm} &\multicolumn{1}{c}{Trajectory}& \multicolumn{2}{c}{Path deviation [rad]}\\
 &\multicolumn{1}{c}{duration [s]}& \hfil mean \hfil &  \hfil max \hfil\\
    \hline
\hspace{0.0002cm} Dataset A \\
\hspace{0.0002cm}\tabitem TOPP-RA &\hfil \textcolor{TABLE_GOOD}{$1.88$} &\hfil \textcolor{TABLE_GOOD}{$<\! 10^{-7}$} &\hfil \textcolor{TABLE_GOOD}{$<\! 10^{-8}$}\\
\hspace{0.0002cm}\tabitem Ruckig pro 0.1 rad&\hfil \textcolor{TABLE_BAD}{$2.13$} &\hfil \textcolor{TABLE_GOOD}{$0.0019$} &\hfil \textcolor{TABLE_GOOD}{$0.0060$}\\
\hspace{0.0002cm}\tabitem Ruckig pro 0.15 rad &\hfil \textcolor{TABLE_GOOD}{$2.02$} &\hfil \textcolor{TABLE_BAD}{$0.0039$} &\hfil \textcolor{TABLE_BAD}{$0.0103$}\\
\hspace{0.0002cm}\tabitem Ruckig pro 0.2 rad&\hfil \textcolor{TABLE_GOOD}{$1.98$} &\hfil \textcolor{TABLE_BAD}{$0.0064$} &\hfil \textcolor{TABLE_BAD}{$0.0153$}\\
\hspace{0.0002cm} Dataset B \\
\hspace{0.0002cm}\tabitem TOPP-RA &\hfil \textcolor{TABLE_BAD}{$7.18$} &\hfil \textcolor{TABLE_GOOD}{$<\! 10^{-6}$} &\hfil \textcolor{TABLE_GOOD}{$<\! 10^{-6}$}\\
\hspace{0.0002cm} Dataset C \\
\hspace{0.0002cm}\tabitem TOPP-RA &\hfil \textcolor{TABLE_BAD}{$7.83$} &\hfil \textcolor{TABLE_GOOD}{$<\! 10^{-5}$} &\hfil \textcolor{TABLE_GOOD}{$<\! 10^{-5}$}\\
    \bottomrule
    \end{tabular*}
    \vspace{0ex}
\label{table:evaluation_dataset_benchmarking}
\end{table}
\begin{table}[t]
     \caption{Impact of the selected jerk limits.}
     \vspace{-0.15cm}
    \makegapedcells
\begin{tabular*}{0.49\textwidth}{@{}p{27mm}p{17mm}p{15mm}p{15mm}} 
    \toprule
\hspace{0.02cm} &\multicolumn{1}{c}{Trajectory}& \multicolumn{2}{c}{Path deviation [rad]}\\
 &\multicolumn{1}{c}{duration [s]}& \hfil mean \hfil &  \hfil max \hfil\\
    \hline
\hspace{0.0002cm} Dataset B \\
\hspace{0.0002cm}\tabitem Jerk limit factor 1.5 &\hfil $6.53$ &\hfil $0.0704$ &\hfil $0.1893$\\
\hspace{0.0002cm}\tabitem Jerk limit factor 2.0 &\hfil $6.48$ &\hfil $0.0546$ &\hfil $0.1614$\\
\hspace{0.0002cm}\tabitem Jerk limit factor 4.0 &\hfil $6.43$ &\hfil $0.0491$ &\hfil $0.1481$\\
\hspace{0.0002cm} Dataset C \\
\hspace{0.0002cm}\tabitem Jerk limit factor 1.5 &\hfil $7.50$ &\hfil $0.0517$ &\hfil $0.1452$\\
\hspace{0.0002cm}\tabitem Jerk limit factor 2.0 &\hfil $7.47$ &\hfil $0.0388$ &\hfil $0.1121$\\
\hspace{0.0002cm}\tabitem Jerk limit factor 4.0 &\hfil $7.38$ &\hfil $0.0253$ &\hfil $0.0874$\\
    \bottomrule
    \end{tabular*}
    \vspace{0ex}
\label{table:dataset_jerk_limit}
\end{table}

In TABLE \ref{table:dataset_jerk_limit}, we finally analyze the impact of the selected jerk limits on the resulting trajectory duration and path deviation. 
As expected, higher jerk limits lead to a faster traversal of the reference paths. 
Moreover, the accuracy of the tracking improves.
Thus, we conclude that higher jerk limits simplify the tracking of a desired reference path length~$u_{\mathrm{ref}}$.

\section{Conclusion and future work}
\label{sec:conclusion}
We presented a method to approximately track multi-dimensional paths considering jerk limits. 
As a first step, we analyzed each path dimension individually.
For every intermediate waypoint of a one-dimensional path, we computed a range of feasible accelerations using a binary search algorithm. 
We then computed a lower and an upper trajectory to achieve minimum and maximum progress on the one-dimensional path, respectively.
This allowed us to control the traversal on the path in such a way that a selected reference path length of a multi-dimensional path could be approximately tracked over time. 
We then applied an iterative scheme, where the reference path length was adjusted at each iteration such that the largest occurring position deviation diminished.
Our evaluation on geometric shapes and datasets with different characteristics showed that our method succeeded in quickly traversing the reference paths while keeping the path deviation low. 

In future work, we would like to further improve our method, e.g., by searching for continuous points in time to switch between the upper and lower trajectory.

\addtolength{\textheight}{-12.2cm}
\bibliographystyle{IEEEtran}
\bibliography{root}

\end{document}